\documentclass[letterpaper, 10 pt, conference]{ieeeconf}  

\IEEEoverridecommandlockouts                              

\usepackage{graphics} 
\usepackage{graphicx}

\usepackage{epsfig} 
\usepackage{amsmath} 
\usepackage{amssymb}  

\usepackage{bm}
\usepackage{mathtools}

\usepackage{algorithm}
\usepackage[noend]{algpseudocode} 
\algnewcommand\AAND{\textbf{ and }}
\algnewcommand\Or{\textbf{ or }}

\usepackage{color}
\usepackage{citesort}
\usepackage{flushend}
\usepackage{url}
\usepackage[breaklinks]{hyperref}
\usepackage[capitalise]{cleveref}
\usepackage{booktabs}
\usepackage{makecell}
\usepackage{multirow}
\usepackage{siunitx}
\usepackage[nolist,nohyperlinks]{acronym}
\acrodef{method}[AOM]{ACRONYM OF METHOD}
\acrodef{gnss}[GNSS]{Global Navigation Satellite System}
\acrodef{ransac}[RANSAC]{Random Sample Consensus}
\acrodef{slam}[SLAM]{Simultaneous Localization And Mapping}
\acrodef{pca}[PCA]{Principal Component Analysis}
\acrodef{ekf}[EKF]{Extended Kalman Filter}
\acrodef{rmse}[RMSE]{Root Mean Square Error} 
\acrodef{ape}[APE]{Absolute Pose Error}
\acrodef{cfar}[CFAR]{Constant False Alarm Rate}
\acrodef{snr}[SNR]{Signal to Noise Ratio}
\acrodef{rcs}[RCS]{Radar Cross Section}
\acrodef{imu}[IMU]{Inertial Measurement Unit}
\acrodef{sgm}[SGM]{Segmi-Global Matching}
\acrodef{dnn}[DNN]{Deep Neural Network}
\acrodef{gru}[GRU]{Gated Recurrent Unit}
\acrodef{hpr}[HPR]{Hidden Point Removal}
\acrodef{raft}[RAFT]{Recurrent All-Pairs Field Transforms}
\acrodef{fov}[FOV]{Field of View}
\acrodef{mclab}[MC-lab]{Marine Cybernetics laboratory}
\acrodef{vio}[VIO]{Visual-Inertial Odometry}
\acrodef{rcm}[RCM]{Refractive Camera Model}
\acrodef{sfm}[SFM]{Structure from Motion}

\newcommand{\Eq}{Eq.}
\newcommand{\pC}{\mathbf{p}_{\camFrame}}  

\newcommand{\ptC}{\mathbf{\tilde{p}}_{\mathcal{C}}}  
\newcommand{\pnC}{\mathbf{\bar{p}}_{\mathcal{C}}}  
\newcommand{\camFrame}{\mathcal{C}}
\newcommand{\bodyFrame}{\mathcal{B}}
\newcommand{\pnCr}{\mathbf{\bar{p}}_{\mathcal{C},r}} 
\newcommand{\pnCl}{\mathbf{\bar{p}}_{\mathcal{C},l}} 
\newcommand{\forwardmap}{\frac{n}{\sqrt{1+r^{2}-n^{2}r^{2}}}}
\newcommand{\inversemap}{\sqrt{n^{2}r_{r}^{2} + n^{2} - r_{r}^{2}}}
\newcommand{\upixel}{\mathbf{u}}
\newcommand{\worldFrame}{\mathcal{W}}
\newcommand{\jac}[2]{\frac{\partial #1}{\partial #2}}
\newcommand{\gl}{\mathbf{g}_{l}}
\newcommand{\gr}{\mathbf{g}_{r}}
\newcommand{\gp}{\mathbf{g}_{p}}

\usepackage{tikz}
\usetikzlibrary{positioning, shapes.geometric, arrows.meta, decorations.pathreplacing, calligraphy}

\DeclareMathAlphabet{\pazocal}{OMS}{zplm}{m}{n}

\newcommand{\Bs}{\pazocal{B}}

\DeclareMathAlphabet{\mathpzc}{OT1}{pzc}{m}{it}

\usepackage{array}
\newcolumntype{C}[1]{>{\centering\arraybackslash}p{#1}}
\newcolumntype{M}[1]{>{\raggedright\arraybackslash}p{#1}}

\usepackage{array} 
\newcolumntype{L}[1]{>{\raggedright\let\newline\\\arraybackslash\hspace{0pt}}m{#1}}	
\newcolumntype{S}[1]{>{\centering\let\newline\\\arraybackslash\hspace{0pt}}m{#1}}
\newcolumntype{R}[1]{>{\raggedleft\let\newline\\\arraybackslash\hspace{0pt}}m{#1}}

\makeatletter
\renewcommand*{\@opargbegintheorem}[3]{\trivlist
  \item[\hskip \labelsep{\itshape #1\ #2}] \textit{(#3)}\ }
\makeatother

\title{\LARGE \bf
Online Refractive Camera Model Calibration in Visual Inertial Odometry
}

\author{Mohit Singh, and Kostas Alexis 
\thanks{This material was supported by the Research Council of Norway Award NO-327292.}
\thanks{The authors are with the Norwegian University of Science and Technology (NTNU), O. S. Bragstads Plass 2D, 7034, Trondheim, Norway {\tt\small mohit.singh@ntnu.no}}
}

\begin{document}

\maketitle
\thispagestyle{empty}
\pagestyle{empty}

\begin{abstract}
This paper presents a general refractive camera model and online co-estimation of odometry and the refractive index of unknown media. This enables operation in diverse and varying refractive fluids, given only the camera calibration in air. The refractive index is estimated online as a state variable of a monocular visual-inertial odometry framework in an iterative formulation using the proposed camera model. The method was verified on data collected using an underwater robot traversing inside a pool. The evaluations demonstrate convergence to the ideal refractive index for water despite significant perturbations in the initialization. Simultaneously, the approach enables on-par visual-inertial odometry performance in refractive media without prior knowledge of the refractive index or requirement of medium-specific camera calibration.

\end{abstract}

\section{Introduction}\label{sec:intro}
Underwater robots are used in various fields including environmental surveillance~\cite{schill2018vertex}, sea floor mapping~\cite{galceran2013planning}, and structural inspection~\cite{hollinger2013active}. To operate autonomously, these systems typically rely on a variety of domain-specific sensors~\cite{wu2019survey}, including $3\textrm{D}$ sonars, acoustics, and Doppler Velocity Log instruments combined with \acp{imu}, while vision cameras are used to observe areas of interest but playing a secondary role in localization~\cite{bahr2009cooperative,paull2013auv,xu2022robust,johannsson2010imaging}. This is due to the fact that visual data in open waters without nearby structures are of limited utility for odometry estimation. However, the growing need for close-up underwater inspection in cluttered settings, such as oil \& gas facilities or shipwrecks, has led to a rising interest in vision-based underwater robots~\cite{shukla2016application,ferrera2019aqualoc,miao2021univio,teixeira2020deep,rahman2019svin2,randall2023flsea}. The affordability of cameras further contributes to this trend. Driven by these observations, research has focused on \ac{vio} techniques underwater~\cite{miao2021univio,ferrera2019aqualoc,joshi2023sm,gu2019environment}. Among the efforts in the domain, it is customary to calibrate the camera model set-up underwater in the area of interest, a choice driven by phenomena such as light refraction in the water and the dependency of refractive index on factors such as temperature, salinity , pressure, wavelength, and more \cite{austinIndexRefractionSeawater1976}.

\begin{figure}[ht]
\centering
    \includegraphics[width=0.99\columnwidth]{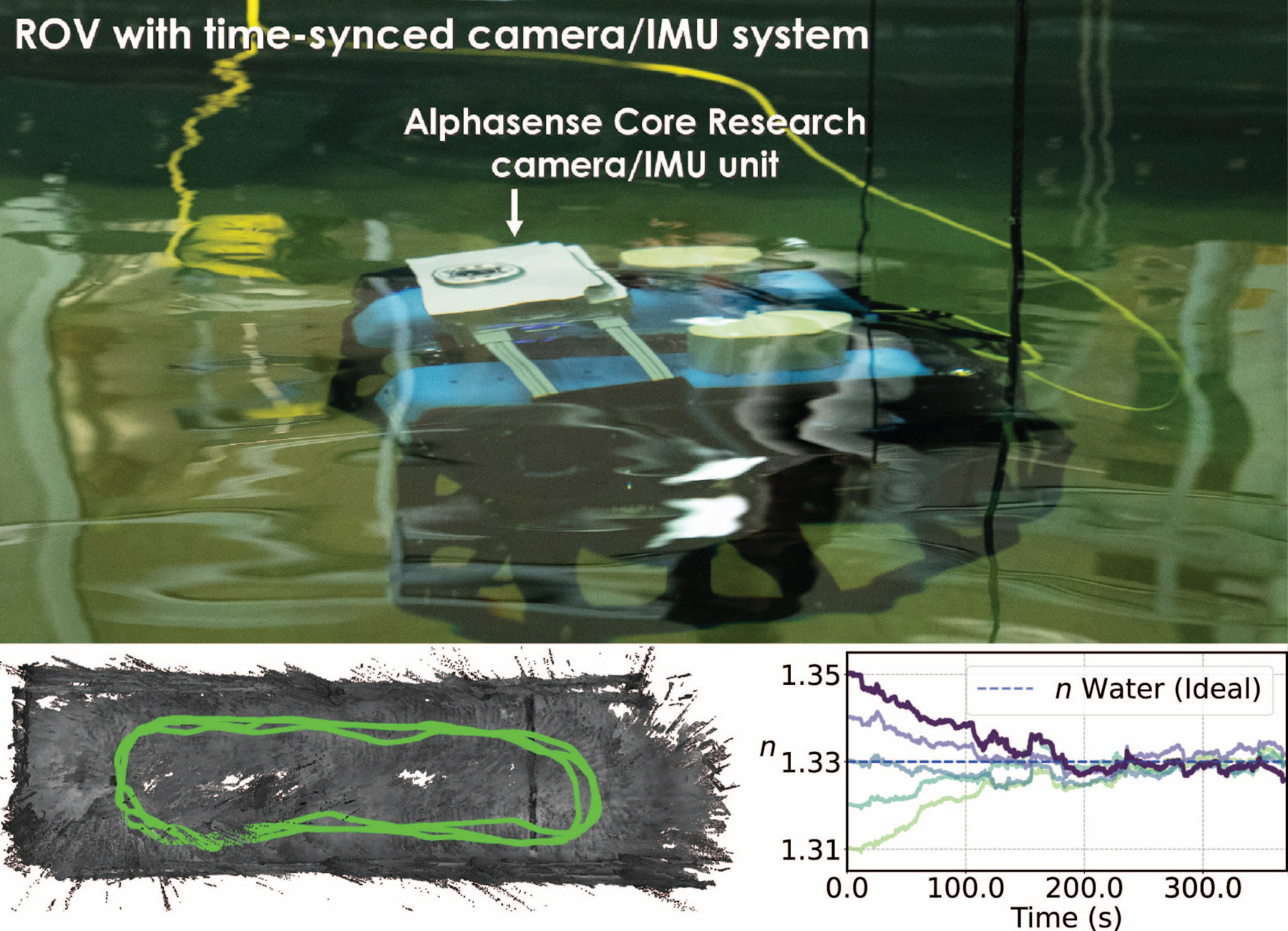}
\vspace{-5ex}
\caption{Instance of the conducted experimental studies employing a Remotely Operated Vehicle integrating a time-synchronized camera/IMU system navigating in a pool subject to diverse light conditions. The proposed approach enables online estimation of the medium's refractive index and thus adjusts the camera model employed within visual-inertial odometry. }
\label{fig:rcm_intro}
\vspace{-4.5ex}
\end{figure}

Motivated by the above and with the goal of eliminating the need for the laborious task of underwater calibration for any camera system with a flat-port, this work contributes \textit{a)} online estimation of a medium's refractive index $n$ as a state variable in a visual-inertial odometry framework for
a general camera-and-IMU system thus enabling versatile underwater \ac{vio}, \textit{b)} online rectification of observed landmarks using the estimated refractive index and camera parameters obtained from conventional calibration in air, \textit{c)} formulation of a sensitivity heuristic for robustness against degenerate motions and noisy image conditions, as well as \textit{d)} verified convergence
of refractive index given a large range of perturbation, highlighting the ability to adapt to a wide range of fluids accessible for robotics and computer vision applications. 

Furthermore, compared to our relevant prior work~\cite{singhOnlineSelfcalibratingRefractive2023}, this contribution \textit{i)} presents a formulation for a monocular camera and \ac{imu} which highlights its generalizability compared to the stereo-only formulation in~\cite{singhOnlineSelfcalibratingRefractive2023}, \textit{ii)} enables the co-estimation of the refractive index embedded into \ac{vio} instead of a 2-step approach where \ac{vio} used a rectified image from a separate refractive module, and \textit{iii)} formulates a sensitivity heuristic as a function of the essential matrix and bearing vector instead of a stereo-specific formulation.  

The presented approach was extensively verified in experimental studies to assess its performance both in terms of real-time refractive index estimation, as well as overall \ac{vio} accuracy and robustness against ground truth. To support this research, an extensive dataset featuring a Remotely Operated Vehicle (ROV) equipped with a time-synchronized 5-camera/\ac{imu} setup in a laboratory pool (Figure~\ref{fig:rcm_intro}) involving diverse light conditions was collected and made publicly available. The collected data are openly released augmenting our previous release in \url{https://github.com/ntnu-arl/underwater-datasets}.

In the remainder of this paper, Section~\ref{sec:related} presents related work, while the proposed camera model is detailed in Section~\ref{sec:cameramodel} and the visual-inertial co-estimation is presented in Section~\ref{sec:underwaterperception}. Experimental studies are shown in Section~\ref{sec:evaluation}, while conclusions are drawn in Section~\ref{sec:concl}.


\section{Related Work}\label{sec:related}
This contribution relates to the body of work on adaptive camera modeling, refractive index estimation and visual odometry across water and other media, alongside the underlying concepts in camera modeling and motion estimation (e.g.,~\cite{hartley2003multiple,fitzgibbon2001simultaneous,barreto2005fundamental,willson1994center}). 
A formulation of the fundamental matrix augmented to account for refractive effects has been detailed in~\cite{chari2009multiple}. 
The study in~\cite{huang2017plate} derives a plate refractive camera model  as a pixel-wise variable viewpoint pinhole model, a caustic surface, a calibration methodology, and a refraction-based triangulation. 
The work in~\cite{treibitz2011flat} concentrates on cameras deviating from the single viewpoint (non-SVP) architecture proposing a physics-grounded model. 
The authors in~\cite{sedlazeck2012perspective} address both perspective and non-perspective camera models and discuss the limitations of the traditional pinhole model underwater.
The work in~\cite{agrawal2012theory} introduces modeling for multiple layer flat refractive geometry and enables refractive index calculation of media with known scene geometry. 
The contribution in~\cite{jordt2013refractive} presents a method for 3D reconstruction from underwater images and a bundle-adjustment technique for autocalibration contingent upon a precise initial estimate.
The authors in~\cite{kawahara2013pixel} approach handling refraction through a pixel-wise varifocal model and use linear extrinsic camera calibration based on a calibrating target. 

Our work also relates to efforts that advanced the application of multiple view geometry in underwater environments. 
In~\cite{chaudhury2015multiple} the authors introduce a camera model to enable underwater scene mapping and compute the medium's refractive index while removing geometric refraction effects based on image point correspondences. 
The work in~\cite{haner2015absolute} focuses on deriving the absolute pose of a camera viewing through a known refractive plane, while emphasizing the intricacies introduced by Snell's law ambiguities. 
The studies in~\cite{hu2023refractive,hu2021absolute} further advance pose estimation in refractive media. 

Recent work has focused on accounting for refractive effects in visual-inertial fusion. 
The study in~\cite{miao2021univio} focuses on improving underwater \ac{vio} by introducing an image rectification technique correcting distortions caused by both water-air refraction and camera lens issues. It employs an approximate SVP model, while calibration is performed underwater. 
Towards self-calibration, the work in~\cite{gu2019environment} presents a method that starts from calibration in air and then estimates environmental indexes in the water under small angle approximations and thus reduced field-of-view cameras. 
Focusing on stereo, \cite{zhang2021open} explicitly considers a refractive camera model but relies on fiducials to enable localization. 
Further on stereo, \cite{servos2013underwater} presents underwater localization and mapping with a refractive camera model calculated offline using calibration images and delivering nonlinear epipolar curves for stereo matching. 
Considering the application-specific alternative of localizing a robot inside the water by structures above the water, \cite{suresh2019through} utilizes an upward-facing stereo camera to build a global ceiling map. 
Outside of this niche set of works that explicitly consider refractive camera models, most approaches in underwater \ac{vio} tend to calibrate their cameras underwater with conventional methods, in shallow waters before being applied in different environments or directly within the target area~\cite{shkurti2011state,joshi2023sm,ferrera2019aqualoc,randall2023flsea,miao2021univio}. Approaches may also be multi-modal (e.g., with the fusion of sonar~\cite{rahman2018sonar} or pressure readings~\cite{hu2022tightly}). 


\section{Refractive Camera Model}\label{sec:cameramodel}
Building upon our previous work in~\cite{singhOnlineSelfcalibratingRefractive2023} this model is derived to enable co-estimation of the refractive index of media (e.g. water) in a Visual-Inertial state estimator. The model applies to general cameras housed in a flat port waterproof casing. It is assumed that the refractive interface, i.e. the glass plate is thin (e.g. \SI{2.5}{\milli\meter}) and the camera lens is negligibly close to the interface (\SI{\sim1}{\milli\meter}).

\subsection{Notations} Below are the notations used in the following sections:
\begin{itemize}
    \item Boldface letters denote matrices or vectors.
    \item $\tilde{(.)}$ denotes the unit vector of the given vector.
    \item $\bar{(.)}$ denotes the normalized vector to a point, normalized by the last dimension of the vector.
    \item $\camFrame$: the camera coordinate frame.
    \item $\bodyFrame$: the IMU (Body) coordinate frame.
    \item $\worldFrame$: the world (inertial) coordinate frame.
    \item $o_\mathcal{C}$: the optical axis
    \item $()_{\diamond , r}$: $r$ in the last subscript denotes that the entity is related to the refractive distortion.
    \item $()_{\diamond , l}$: $l$ in the last subscript denotes that the entity is related to the lens distortion.
\end{itemize}

\begin{figure}[ht]
    \centering
    \includegraphics[width=0.70\linewidth]{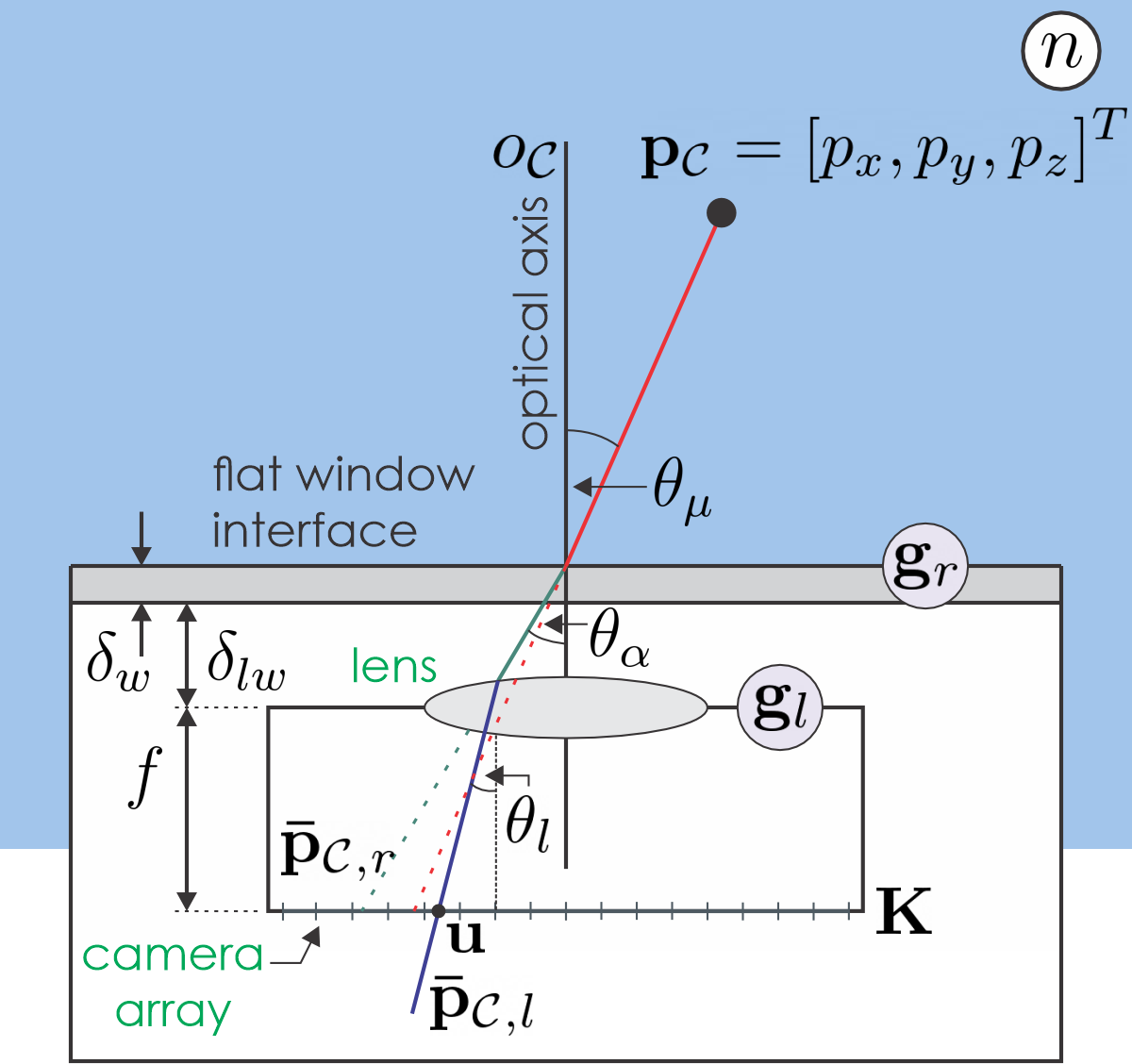}
    \caption{Visualization of a ray from a point $\pC$ in the refractive media (e.g. water), undergoing refractive distortion modelled by $\gr$, followed by lens distortion modelled by $\gl$. Lastly, the point is projected onto camera array, modelled by $\mathbf{K}$, at pixel coordinate $\upixel$. The thickness of the refractive interface $\delta_{w}$ and the distance of the interface from the camera lens $\delta_{lw}$ are assumed to be small.}
    \label{fig:camera_geom_2d}
\end{figure}

\subsection{Refractive distortion}
The incidence ray from a point $\pC=[p_{x}, p_{y}, p_{z}]^{T}\in \mathbb{R}^3$ in a medium (with refractive index $n$), expressed in the camera coordinate frame $\camFrame$, makes an angle $\theta_{\mu}$ from the camera optical axis $o_{\mathcal{C}}=[0,0,1]^{T}$. This ray undergoes refraction upon entering the camera case through the flat-port and is deflected, making an angle $\theta_{\alpha}$ from $o_{\camFrame}$ after refraction (Figure~\ref{fig:camera_geom_2d}). These angles are related by the Snell's law as

\begin{equation}
    \sin{\theta_{\alpha}}=n\sin{\theta_{\mu}}
\label{eq: snell's law}
\end{equation}
We write the projection of the point on a plane at a unit distance as
$\pnC =\mathbf{g}_p(\pC)$. It relates $\mathbf{p}_\mathcal{C}$ to a projected coordinate frame such that

\begin{equation}
\Bar{\mathbf{p}}_\mathcal{C} = \left[\bar{p}_x, \bar{p}_y \right]^{T}= \left[\frac{p_x}{p_z}, \frac{p_y}{p_z} \right]^{T}
\end{equation}
Let $r=\sqrt{p_{x}^{2}+p_{y}^{2}}$, then $\sin\theta_{\mu}$ can be expressed as

\begin{equation}
    \sin\theta_{\mu} = \frac{r}{\sqrt{1+r^{2}}}
\end{equation}
Likewise, we can write for the refracted ray with normalized coordinates as $\pnCr = [\bar{p}_{x,r}, \bar{p}_{y,r}]^{T}$ expressing $\sin\theta_{\alpha}$

\begin{equation}
    \sin\theta_{\alpha} = \frac{r_{r}}{\sqrt{1+r_{r}^{2}}}
\end{equation}
Using equation (\ref{eq: snell's law}) we can write

\begin{equation}
     \frac{r_{r}}{\sqrt{1+r_{r}^{2}}}=n\frac{r}{\sqrt{1+r^{2}}}
\label{eq: m derivation}
\end{equation}
Let $m=r_{r}/r$ then by rearranging Eq.~(\ref{eq: m derivation}) we can write $m$ in terms of $n$ and $r$ as

\begin{equation}
    m(n, r) = \forwardmap
\end{equation}
and also in terms of $n$ and $r_{r}$ as
\begin{equation}
    m(n, r_{r}) = \inversemap
\end{equation}
This allows us to derive the forward (distortion) and inverse (undistortion) maps.

\subsubsection{Forward Mapping}
Accordingly, the forward (distortion) mapping $\pnC\to \pnCr$ takes the form

\begin{subequations}
\begin{align}
\pnCr &= m(n,r)\pnC\\
r &= \sqrt{\bar{p}_{x}^{2} + \bar{p}_{y}^{2}}
\end{align}
\label{forward mapping}
\end{subequations}
\subsubsection{Inverse Mapping}
Similarly, the inverse (undistortion) mapping can be written as  $\pnCr\to \pnC$

\begin{subequations}
\begin{align}
\pnC&=\frac{\pnCr}{m(n,r_{r})}\\
r_{r} &= \sqrt{\bar{p}_{x,r}^{2} + \bar{p}_{y,r}^{2}}
\end{align}
\label{eq: inverse mapping}
\end{subequations}

\subsection{Derivations for Iterative Estimation}
For integration into \ac{vio}, we first need to model the projection of a point $\pC\in\mathbb{R}^{3}$ onto the pixel coordinates $\upixel\in\mathbb{R}^{2}$ by incorporating both refractive and lens distortions. A combined model for this can be written as

\begin{equation}
  \upixel = \mathbf{K}\mathbf{g}_l(\mathbf{g}_{r}(n,\mathbf{g}_{p}( \pC))) 
\label{eq: total model}
\end{equation}
where, $\mathbf{K}$ is the camera intrinsics matrix, consisting of focal lengths $(f_{x},f_{y})$ and image center $(c_x,c_y)$, represented as

\begin{equation}
    \mathbf{K} = \begin{bmatrix}
f_{x} & 0 & c_{x} \\
0 & f_{y} & c_{y} \\
0 & 0 & 1
\end{bmatrix}~\textrm{,}
\label{eq: camera matrix}
\end{equation}
$\gl$ is the lens distortion function, $\gr$ is the refractive distortion function, and 
 $\gp$ is the function that models the projection on a plane at a unit distance. For brevity in the partial derivatives, we drop the function arguments.
The partial derivatives of $\upixel$ (\Eq~(\ref{eq: total model})) w.r.t. (with respect to) $\pC$ and $n$ are given by:

\begin{subequations}
\begin{align}
\jac{\upixel}{\pC} &= \mathbf{K}_{f} \jac{\gl}{\pnCr} 
                                \jac{\gr}{\pnC}
                                \jac{\gp}{\pC}\\   
\jac{\upixel}{n}&= \mathbf{K}_{f} \jac{\gl}{\pnCr} 
                                \jac{\gr}{n}
\end{align}
\label{eq: total model jacobian with pC}
\end{subequations}
where $\mathbf{K}_{f}=\mathit{diag}\left [f_{x}~ f_{y} \right ]$ (diagonal matrix). The terms of chain-rule (partial derivatives) along with their function definitions are described below, which are used later in an IEKF for \ac{vio} as in~\cite{bloeschIteratedExtendedKalman2017}. 

\subsubsection{Projection Function}
$\pnC =\mathbf{g}_{p}(\pC)$ 

The Jacobian w.r.t. $\mathbf{p}_\mathcal{C}$ takes the form

\begin{equation}
\jac{\gp}{\pC} = \begin{bmatrix}
    \frac{1}{p_z}  & 0 & \frac{-p_x}{p_{z}^{2}} \\
    0  & \frac{1}{p_z} & \frac{-p_y}{p_{z}^{2}}
\end{bmatrix}
\label{eq: projection jacobian}
\end{equation}

\subsubsection{Refractive Distortion Function}
$\pnCr = \gr(n, \pnC)$

\small
\begin{equation}
\pnCr=m\left( n, \sqrt{\bar{p}_{x}^{2}+\bar{p}_{y}^{2}}\right)\pnC
\label{eq: refractive distortion}
\end{equation}
\normalsize

The Jacobian w.r.t. $n$ takes the form

\begin{equation}
\jac{\gr}{n} = \begin{bmatrix}
\frac{\Bar{p}_{x} \left(\sqrt{h} n^{2} r^{2} + \sqrt{h}^{3}\right)}{h^{2}}\\ \\
\frac{\Bar{p}_{y} \left(\sqrt{h} n^{2} r^{2} + \sqrt{h}^{3}\right)}{h^{2}}
\end{bmatrix}
\label{eq: refractive jacobian wrt n}    
\end{equation}
where $h=\left(1 + r^{2}- n^{2} r^2\right)$.

The Jacobian w.r.t. $\pnC$ takes the form

\small
\begin{equation}
\jac{\gr}{\pnC} = \begin{bmatrix}
\frac{n \left( \sqrt{h} \Bar{p}_{x}^{2} \left(n^{2} - 1\right) + \sqrt{h}^{3}\right)}{h^{2}} & \frac{n \Bar{p}_{x}\Bar{p}_{y} \left(n^{2} - 1\right)}{\sqrt{h}^{3}}\\ \\
\frac{n \Bar{p}_{x} \Bar{p}_{y} \left(n^{2} - 1\right)}{\sqrt{h}^{3}} &\frac{n \left( \sqrt{h} \Bar{p}_{y}^{2} \left(n^{2} - 1\right) + \sqrt{h}^{3}\right)}{h^{2}}
\end{bmatrix}
\label{eq: refractive jacobian wrt pnC}
\end{equation}
\normalsize

\subsubsection{Lens Distortion Function} $\pnCl = \gl(\pnCr)$

In this work we use the equidistant model \cite{kannalaGenericCameraModel2006} which is defined as

\small
\begin{subequations}
\begin{align}
    \pnCl&=\frac{\theta_{e}}{r_{r}}\pnCr\\
    \theta_{e} &= \theta (1+k_1\theta^{2} + k_2\theta^{4} + k_3\theta^{6} + k_4\theta^{8}) \\
    \theta &= \tan^{-1}{(r_{r})}
\end{align}
\end{subequations}
\normalsize
where $r_{r}$ is same as in Eq. (\ref{eq: inverse mapping}), $k_1, k_2, k_3, k_4$ are the equidistant model parameters and $\theta_{l}=\tan^{-1}(\theta_{e})$ in Figure~\ref{fig:camera_geom_2d}.

The Jacobians are then given as

\small
\begin{equation}
    \jac{\gl}{\pnCr} = \jac{\pnCl}{\pnCr} + \jac{\pnCl}{r_{r}}\jac{r_{r}}{\pnCr} + \jac{\pnCl}{\theta_{e}}\jac{\theta_{e}}{\theta}\jac{\theta}{r_{r}}\jac{r_{r}}{\pnCr}
\label{eq: lens distortion jacobian wrt pnCr}
\end{equation}
\normalsize
where

\small
\begin{subequations}
\begin{align}
    \jac{\pnCl}{\pnCr} &= \mathit{diag}\begin{bmatrix}
        \frac{\theta_{e}}{r_{r}} & \frac{\theta_{e}}{r_{r}}
    \end{bmatrix}\\
    \jac{\pnCl}{r_{r}}&=-\frac{\theta_{e}}{r_{r}^{2}}\pnCr
    \\
    \jac{r_{r}}{\pnCr}&=\frac{1}{r_{r}}\pnCr^{T}\\
    \jac{\pnCl}{\theta_{e}}&=\frac{1}{r_{r}}\pnCr\\
    \jac{\theta_{e}}{\theta}&= 1+3k_{1}\theta^2+5k_{2}\theta^4+7k_{3}\theta^6+9k_{4}\theta^{8}\\
    \jac{\theta}{r_{r}}&=\frac{1}{r_{r}^{2}+1}
\end{align}
\end{subequations}
\normalsize

 The above derivations allow us to use an iterative framework for estimation of the refractive index and to project the points in the world to pixel coordinates and vice-versa. These derivations are not specific to a particular Visual-Inertial Odometry framework and can be applied to a suitable method.

\section{Refractive Visual-Inertial Odometry}\label{sec:underwaterperception}
The novel refractive camera model tailored to online estimation was integrated into a state-of-the-art \ac{vio} system, namely ROVIO~\cite{bloeschIteratedExtendedKalman2017}, to develop a resilient solution for underwater localization in diverse media without any assumption of knowledge of the exact refractive index or need to laboriously calibrate the camera/\ac{imu} set-up underwater. The choice of ROVIO was motivated by its robust formulation, delay-free initialization, and good low-light performance as evaluated in \cite{cerberus_science,cerberus_finals,cerberus_tunnel_urbam}. It uses multi-level image patches as a frontend and provides the constraints based on photometric error. Its backend is an Iterated Extended Kalman Filter (IEKF). ROVIO uses patchwise QR-decomposition, reducing the error dimensionality and ensuring computational efficiency in the IEKF update step. 

We incorporate the estimation of refractive index $n$ in the direct image intensity errors and accordingly re-derive the filter innovation term. ROVIO employs a robocentric formulation thus estimating landmarks relative to the pose of the camera. Estimated landmarks are decomposed and expressed in terms of a bearing vector and an inverse depth parametrization. For the filter formulation, we consider $\bodyFrame$ as the \ac{imu}-fixed coordinate frame, $\camFrame$ as the camera-fixed frame, and $\worldFrame$ as the world (inertial) frame, while the resulting state vector and associated covariance are denoted as $\mathbf{s}$ and $\mathbf{\Sigma}$, respectively. The method can support multi-camera systems, but analogous to the original work, here we focus on the general formulation for the monocular case. Note that $\ptC$ denotes the unit vector to the point $\pC$, following the representation of 3D unit vectors in \cite{bloeschIteratedExtendedKalman2017}. The state vector augmented with the refractive index $n$ is then given by

\begin{equation}
    \mathbf{s} = \left [ \mathbf{r}~\mathbf{q}~\mathbf{v}~\mathbf{b}_f~\mathbf{b}_\omega~\mathbf{c}~\mathbf{z}~| n |~ \tilde{\mathbf{p}}_{\camFrame,0},...,\tilde{\mathbf{p}}_{\camFrame,J}~\rho_0,...,\rho_J \right]
\end{equation}

where $\mathbf{r},\mathbf{v}$ denote the robocentric position and velocity of the \ac{imu} expressed in $\bodyFrame$, $\mathbf{q}$ is the \ac{imu} attitude (map from $\bodyFrame\rightarrow \worldFrame$), $\mathbf{b}_f,\mathbf{b}_\omega$ are the bias of accelerometer and gyroscope expressed in $\Bs$, $\mathbf{c},\mathbf{z}$ denote the translational and rotational components of the camera extrinsics against the \ac{imu} (maps from $\bodyFrame\rightarrow \camFrame$), $n$ is modeled as $\dot{n}=\boldsymbol{\mathcal{M}_{n}}$, where $\boldsymbol{\mathcal{M}_{n}}$ is the process noise, $\tilde{\mathbf{p}}_{\camFrame,j}$ is the bearing vector to the $j$-th feature (out of maximum $J$ features) expressed in $\camFrame$ and $\rho_{j}$ is the associated depth parameter under the parametrization $d(\rho_{j}) = 1/\rho_{j}$ for the feature distance $d_{j}$. Although ROVIO can support extrinsic estimation, in the results presented in this work the extrinsics $\mathbf{c},\mathbf{z}$ are constant and set to the values estimated offline through Camera-IMU calibration in air. The interested reader may refer to ROVIO~\cite{bloeschIteratedExtendedKalman2017} for a more detailed formulation of it.

\subsection{Projection model and linear Warping}
Given known camera calibration in air, $\pC$ can be mapped to $\upixel$ for some $n$ using Eq.~(\ref{eq: total model}), denoted as $\upixel=\pi(n, \ptC)$. Also, given the inverse of refractive distortion Eq.~(\ref{eq: inverse mapping}) and lens distortion, the inverse projection can be written as ${\ptC}=\pi^{-1}(n, {\upixel})$. A linear warping matrix  $\boldsymbol{D}$ then accounts for the change in perspective of the patch, while also undergoing distortion in consecutive frames. For two image frames $(1,2)$ this can be given by stacking the Jacobians of the following functions: inverse projection $\tilde{\mathbf{p}}_{\camFrame,1}=\pi^{-1}(n, {\upixel}_{1})$, process model $\tilde{\mathbf{p}}_{\camFrame,2}=f(\tilde{\mathbf{p}}_{\camFrame,1})$, and projection ${\upixel}_{2}=\pi(n, \tilde{\mathbf{p}}_{\camFrame,2})$.

Then, the linear warping can be written as

\small
\begin{equation}
    \boldsymbol{D}=\jac{\pi(n, \tilde{\mathbf{p}}_{\camFrame,2})}{\tilde{\mathbf{p}}_{\camFrame,2}}\jac{f({\tilde{\mathbf{p}}_{\camFrame,1})}}{\tilde{\mathbf{p}}_{\camFrame,1}}\jac{\pi^{-1}(n, {\upixel}_{1})}{{\upixel}_{1}} \in \mathbb{R}^{2\times2}
\label{eq: linear warping}
\end{equation}
\normalsize

\subsection{Photometric error}
The photometric error of patch $j$ at image pyramid level $l$ $(e_{l, j})$ takes the same form as in ROVIO:

\small
\begin{equation}
    e_{l, j}(\upixel, P, I, \boldsymbol{D})=P_{l}(\upixel_{j}) - aI_{l}({\upixel}_{l}c_{l}+\boldsymbol{D}{\upixel}_{j}) - b
\label{eq: photometric error}
\end{equation}
\normalsize
where $P$ is the multi-level image patch centered at $\upixel$, $I$ is the input image, $a$ and $b$ are the scalars to account for illumination variation in consecutive images, $l\in\{0,\cdots, L\}$ is the patch level and entities with $l$ in subscript denote the computation at that level of the image pyramid, while $c_l=0.5^{l}$ is the factor to scale the error based on the pyramid level. The linearized error equation at patch location estimate $\hat{\upixel}$ can be written as

\small
\begin{equation}
    \boldsymbol{e}(\hat{\upixel}+\delta\upixel, P, I, \boldsymbol{D})=\boldsymbol{J}(\hat{\upixel}, I, \boldsymbol{D})\delta\upixel + \boldsymbol{e}(\hat{\upixel}, P, I, \boldsymbol{D})
\label{eq: stacked errors}
\end{equation}
\normalsize
where $\boldsymbol{e}$ is the stacked errors from Eq.~(\ref{eq: photometric error}), $\boldsymbol{J}$ denotes the corresponding Jacobian and $\delta\upixel$ is the correction term. Further, the normal equation can be given by

\small
\begin{equation}
    \boldsymbol{J}(\hat{\upixel}, I, \boldsymbol{D})^{T}\boldsymbol{J}(\hat{\upixel}, I, \boldsymbol{D})\delta\upixel=-\boldsymbol{J}(\hat{\upixel}, I, \boldsymbol{D})^{T}\boldsymbol{e}(\hat{\upixel}, P, I, \boldsymbol{D}) 
\label{eq: normal equation}
\end{equation}
\normalsize

To employ dimensionality reduction of the error and its Jacobian, QR decomposition is conducted 

\scriptsize
\begin{equation}
    \boldsymbol{J}({\upixel}_{j},I,\boldsymbol{D}_{j})=\begin{bmatrix}
        \boldsymbol{Q}_{1}({\upixel}_{j}, I, \boldsymbol{D}) & \boldsymbol{Q}_{2}({\upixel}_{j}, I, \boldsymbol{D})
    \end{bmatrix}
    \begin{bmatrix}
        \boldsymbol{R}_{1}({\upixel}_{j}, I, \boldsymbol{D})\\
        \boldsymbol{0}
    \end{bmatrix}
\label{eq: QR decomposition}
\end{equation}
\normalsize
where $\boldsymbol{R}_{1}({\upixel}_{j}, I, \boldsymbol{D})$ corresponds to the upper triangular matrix that has full row-rank $2$ for distinct corner features, row-rank $1$ for line features and $\boldsymbol{Q}_{1}$ and $\boldsymbol{Q}_{2}$ have orthogonal columns. 

\subsection{Innovation Term}

The innovation term $\boldsymbol{y}_{i,j}$ with the projection function incorporating the refractive camera model at $i_{th}$ iteration and $j_{th}$ patch, $\hat{\upixel}=\pi(n^{+}, \tilde{\mathbf{p}}_{\camFrame~i,j}^{+} )$ can be written as 

\begin{equation}
    \boldsymbol{y}_{i,j}=\boldsymbol{Q}_{1}(\pi(n^{+}, {\pC}_{i,j}^{+}), I, \boldsymbol{D}_{j})^{T}\boldsymbol{e}(\pi(n^{+}, {\pC}_{i,j}^{+}), P_{j}, I, \boldsymbol{D}_{j})
\label{eq: innovation term}
\end{equation}
\normalsize
where $(\cdot)^{+}$ is the a-posteriori estimate. Then $\boldsymbol{H}_{i,j}$ can be written as Jacobian of the decomposed innovation term w.r.t $\pC$ and $n$ respectively as

\small
\begin{equation}
    \boldsymbol{H}_{i,j}(\pC)=\boldsymbol{R}_{1}(\pi(n^{+}, {\pC}_{i,j}^{+}), I, \boldsymbol{D}_{j})\jac{\pi}{{\pC}}(n^{+}, {\pC}_{i,j}^{+})
\label{eq: innovation Jacobian with mu}
\end{equation}
\begin{equation}
    \boldsymbol{H}_{i,j}(n)=\boldsymbol{R}_{1}(\pi(n^{+}, {\pC}_{i,j}^{+}), I, \boldsymbol{D}_{j})\jac{\pi}{{n}}(n^{+}, {\pC}_{i,j}^{+})
\label{eq: innovation Jacobian with n}
\end{equation}
\normalsize

\subsection{Sensitivity Heuristic}
Refractive index estimation is vulnerable to sub-pixel errors and noisy tracking of landmarks. Two conditions where the signal-to-noise ratio deteriorates are if the points move along the radial or tangential direction from the image center. Motivated by this, we develop a heuristic map to scale Jacobians which reduces the effect of points near the degraded regions.

Let, $\boldsymbol{T}_{\bodyFrame_{t+1}, \bodyFrame_{t}}\in \mathcal{SE}(3)$ be the rigid transformation matrix from $\bodyFrame$ at time $t$ to $\bodyFrame$ at time $t+1$. Then, the rigid transformation from $\camFrame$ at $t$ to $t+1$ is $\boldsymbol{T}_{\camFrame_{t+1}, \camFrame_{t}} = \boldsymbol{T}_{\camFrame_{t+1}, \bodyFrame_{t+1}}\boldsymbol{T}_{\bodyFrame_{t+1}, \bodyFrame_{t}}\boldsymbol{T}_{\bodyFrame_{t}, \camFrame_{t}}\in \mathcal{SE}(3)$, where $\boldsymbol{T}_{\camFrame_{t+1}, \bodyFrame_{t+1}}=\boldsymbol{T}_{\bodyFrame_{t}, \camFrame_{t}}^{-1}$ denotes the camera-IMU extrinsics.

Further, assuming a small rotation in frame-to-frame motion, the Epipolar line can approximate the motion of the landmark in image coordinates in consecutive frames. The essential matrix \cite{hartley2003multiple} is $\mathbf{E}=[\mathbf{t}]^{\times}\mathbf{R}$ where $\mathbf{R}\in\mathit{SO}(3)$ is the rotation matrix and $\mathbf{t}\in\mathbb{R}^{3}$ is the translation from the homogeneous transformation $\boldsymbol{T}_{\camFrame_{t+1}, \camFrame_{t}}$ and the epipolar line in normalized coordinates is $\boldsymbol{\lambda}=\mathbf{E}^{T}\tilde{\mathbf{p}}_{\camFrame,j}\in \mathbb{R}^{3}$. Let $\tilde{\boldsymbol{\lambda}}\in \mathbb{R}^{2}$ be the unit vector along the epipolar line in image coordinates, and $\tilde{\mathbf{r}}$ be the unit vector of the line joining the image center and pixel $\upixel$. Then $\theta=\cos^{-1}(\tilde{\boldsymbol{\lambda}}\cdot\tilde{\mathbf{r}})$ is the angle between the epipolar line and the radial line. The heuristic $v$ can then be given by

\small
\begin{equation}
    v={|\sin(2\theta)|}^{q}r^{k}
\end{equation}
\normalsize
where $0<q\leq 1$ and $0<k\leq1$ are tunable scalars. We then apply this heuristic patch and level-wise in the \ac{vio} estimator by using the scaled Jacobian $v_{i,j}\boldsymbol{H}_{i,j}(n)$ in place of the non-weighted form in Eq.~(\ref{eq: innovation Jacobian with n}). In the following results we use $q=0.5$ and $k=0.8$.


\section{Evaluation Studies}\label{sec:evaluation}
A host of experiments were conducted to assess the performance of the proposed general refractive camera model and online co-estimation of refractive index and odometry. During the presented studies, we investigate convergence of the estimated refractive index and accompanying odometry results. We initialize the augmented \ac{vio} with refractive index values that represent significant changes of media versus the water in which the tests took place. For a general perspective, for liquids at $20^\circ$C, water has a refractive index of $1.33$ (wavelength of $587.6\textrm{nm}$)~\cite{daimon2007measurement}, acetone's value is $1.36$, $60\%$ glucose solution in water has $1.44$~\cite{lide2004crc} and benzene has $1.5$. Saline water also assumes values higher than $1.33$, e.g. $1.35$ for NaCl mass fraction $wt\%$ of $10$, wavelength of $589\textrm{nm}$ and under certain temperatures. Water approaching $100^\circ$C reaches a refractive index of almost $1.31$. 

\subsection{Robot Experimental Set-up}
To verify the proposed contributions, we use an underwater robot (Figure~\ref{fig:rcm_intro}) described in~\cite{singhOnlineSelfcalibratingRefractive2023} that is based on the BlueROV remotely operated vehicle and integrates an Alphasense Core Research Development kit and NVIDIA Orin AGX. The results employ its left forward facing camera with a \ac{fov} $D\times H\times V = 165.4^\circ \times 126^\circ \times 92.4^\circ$ and a focal length of $2.4\textrm{mm}$ measured in air, alongside the time-synchronized BMI085 \ac{imu}. The camera is mounted on the top of the robot and is inclined down by $16^\circ$.

\begin{figure*}[ht!]
\centering
    \includegraphics[width=0.99\textwidth]{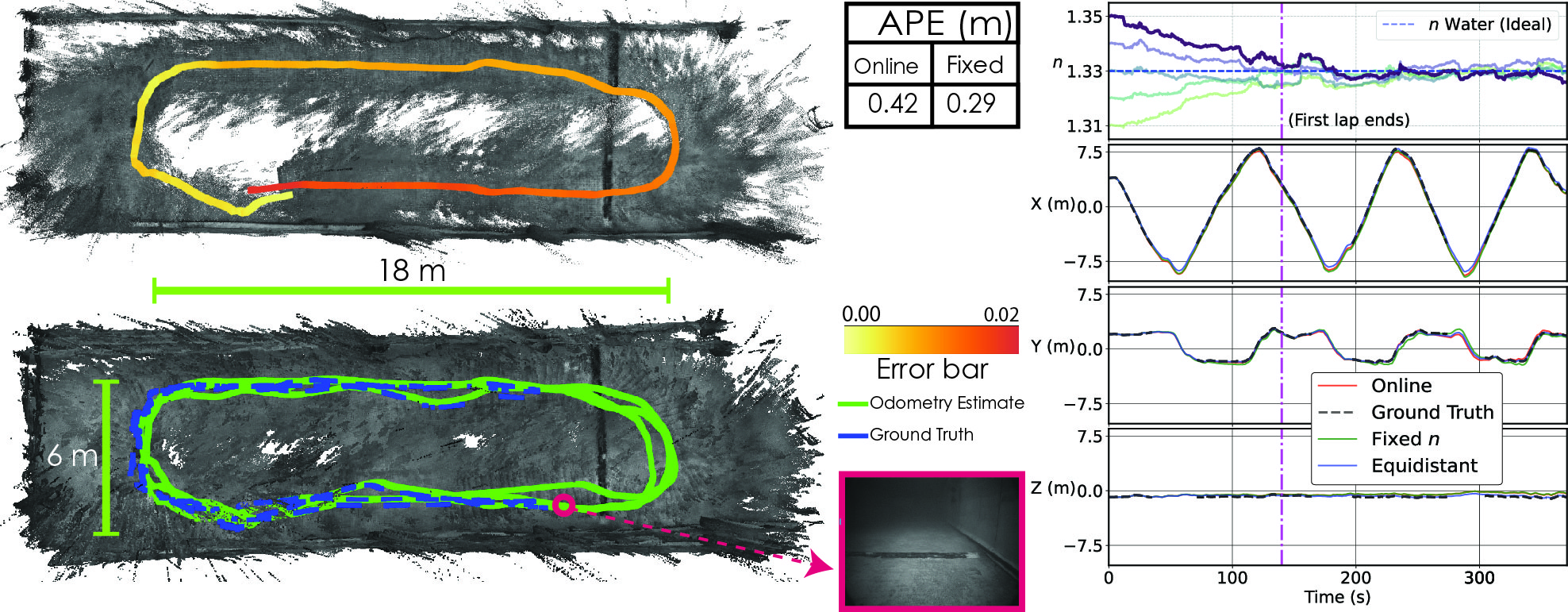}
\vspace{-2ex} 
\caption{Detailed evaluation on a rectangular trajectory with good ambient light conditions (Trajectory 1). Top Left: The first lap of the estimated trajectory, given refractive index $n$ initialization of $1.35$, with path colorized based on the refractive index absolute error from the ideal value of the refractive index of water $n=1.33$. The bottom left plot shows the full trajectory for $370~\textrm{sec}$ along with the ground truth (dashed blue). Top right: the plot showing refractive index $n$ vs. time with initialization of $n$ varying from $1.31$ to $1.35$. Bottom right: The comparison of odometry against ground truth vs. time for online estimation with initialization from $n$ equal to $1.35$, fixed value of $n$ at $1.33$, and calibration of the camera directly inside the water of the same pool using an equidistant model is water. The map from the accumulated point cloud is generated using \cite{lipson2021raftstereo} for visualization.}
\label{fig:detailedresult}
\vspace{-2ex} 
\end{figure*}

\subsection{Dataset for Refractive Visual-Inertial Odometry}

The aforementioned robotic system was deployed inside the \ac{mclab} of NTNU which offers a water tank (dimensions: $L \times B \times D = 40 \textrm{m} \times 6.45\textrm{m} \times 1.5 \textrm{m}$) that is partially supported with Motion Capture (MoCap). Specifically, we utilize the MoCap capabilities above the water by installing a pole and markers on the robot that allow it to be tracked. The MoCap is based on a Qualisys solution and offers partial coverage due to structural occlusions. During the conducted trajectories, all the $5$ cameras and \ac{imu} of the Alphasense Core Research Development kit are recorded, but as described only the left forward-facing camera is employed for this work. Nevertheless, the complete dataset is released to the community by augmenting our previous release found at \url{https://github.com/ntnu-arl/underwater-datasets}.

In further detail, a total of $6$ trajectories were first conducted and organized based on $3$ motion patterns and $2$ ambient light condition, namely either with ceiling lights on or off. Additionally, one more trajectory was collected to demonstrate the convergence of refractive index is tested under ``wild'' initial guess. Every mission begins with the robot being approximately at the center of the tank (along the length) and covers of three laps, while the camera always points in the direction of motion. In the trajectories of the first motion pattern, the robot follows a rectangular shape. In the trajectories of the second motion pattern, the robot is piloted in a figure-$8$ pattern. In the third motion pattern, the robot moves along the length of the path while primarily keeping close the center width-wise. The motivation behind the motion patterns is driven by: $i)$ gradually increasing the distance to visual surfaces seen by the robot, and $ii)$ gradually introducing more variations in the motion. Thus, the rectangular motion patterns allow close proximity to walls, and more reliable visual features are tracked. The figure $8$ and the last motion patterns further increase the average distance to visual features and introduce more varying motions.

\subsection{Detailed Evaluation on Single Trajectory}
We present the trajectory $1$, meaning a rectangular pattern in good ambient light conditions, in detail to highlight the convergence of the refractive index estimated by the proposed method. The results in Figure~\ref{fig:detailedresult} show the trajectory with a colorized path. We compare it against the odometry estimated with the same camera model initialized and kept fixed at $n=1.33$, alongside the odometry results using an equidistant camera model calibrated inside the water of this exact pool. As shown, our method enables refractive index estimation and robust odometry without the need of calibration inside the water and despite the initialization from significantly different refractive index value. 

\subsection{Collective Evaluation Studies}
\begin{figure}[ht!]
\centering
    \includegraphics[width=0.99\columnwidth]{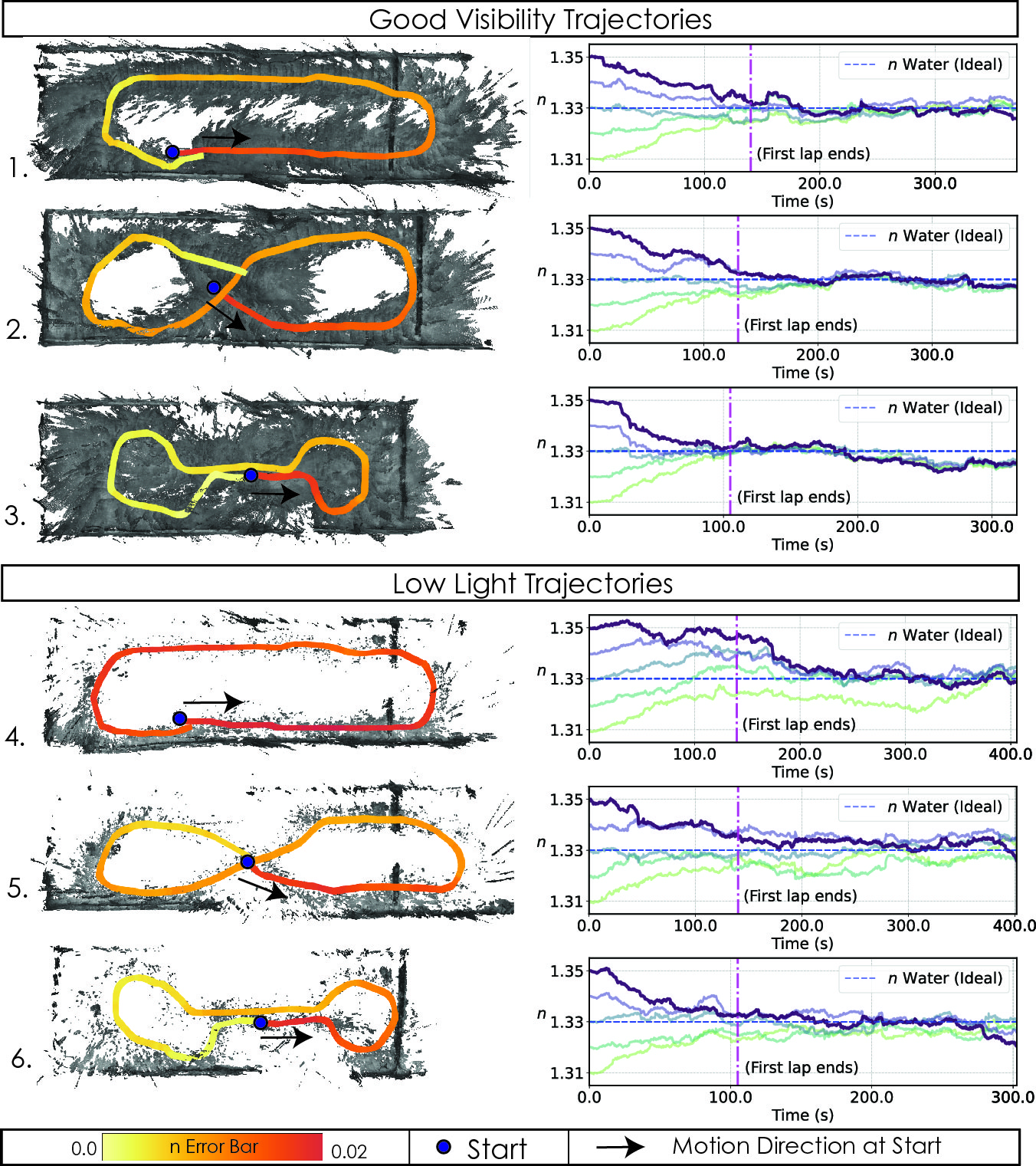}
\caption{The top-down plots of trajectory on left show the colorized path based on the absolute difference from the ideal value of the refractive index for water $n=1.33$ when initialized with $1.35$ (purple line in the right plot starting from $1.35$). The path is colored according to the error bar shown at the bottom. The blue circle is the starting point of the robot. For clarity of visualization, only the first lap of motion is shown. The refractive index vs. time plots on the right show, for the complete mission, the convergence of refractive index $n$ from perturbations ranging from $1.31$ to $1.35$. The lack of visual texture results in the sparseness of the point cloud generated using~\cite{lipson2021raftstereo} for low-light conditions.}
\label{fig:collective_figure}
\end{figure}

\begin{figure}[ht!]
\centering
    \includegraphics[width=0.99\columnwidth]{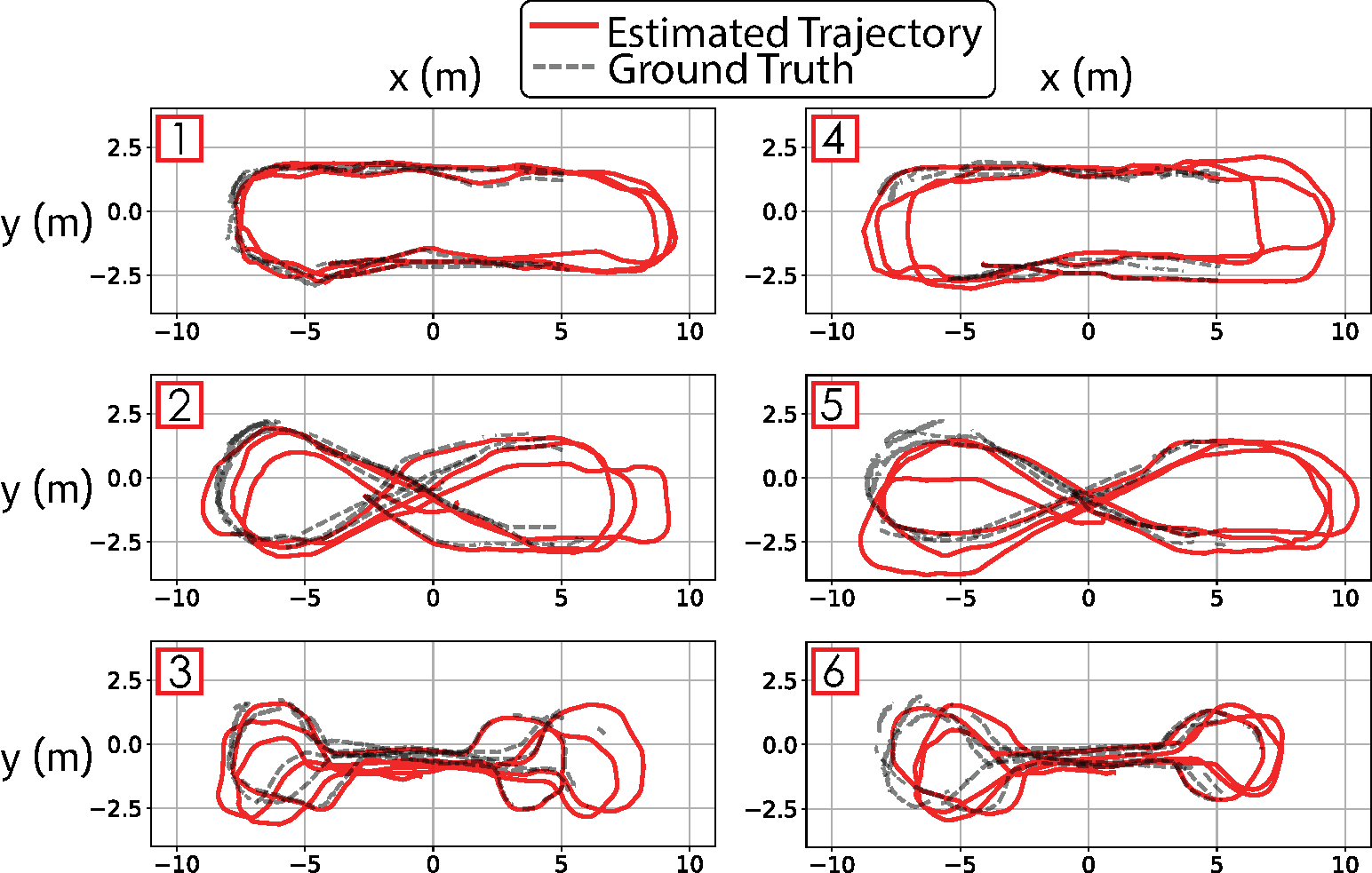}
\caption{Collective plots for trajectories comparing against ground truth for an initial $n$ of $1.35$. On the left, odometry estimates for trajectories 1-3 (good visibility). Similarly, the plots on the right show the estimates in the low-light trajectories 4-6.}
\label{fig:collective_figure_traj}
\end{figure}

In this evaluation, we plot the refractive index vs. time plots for all collected trajectories to highlight the convergence of the refractive index given perturbed initial conditions that range from $1.31$ to $1.35$. Figure~\ref{fig:collective_figure} presents the associated results of odometry for the first loop of the motion (given initial $n$ of $1.35$) alongside the mission-complete plots of $n$ convergence for multiple initial conditions. Figure~\ref{fig:collective_figure_traj} presents odometry for each mission in full (given initial $n$ of $1.35$). The choice of perturbation for the refractive index aims to assess the convergence of refractive index values that cover the range of what water exhibits in very different environmental conditions, thus demonstrating the ability to use the method with no specific knowledge of the medium's refractive index. For trajectories $1-3$, good visibility allows the refractive index to converge within $\pm0.005$ within $150~
\textrm{sec}$ (approximately one loop of the conducted paths). For the low-light trajectories, the convergence is delayed due to a lack of abundant visual features. The lack of visual texture is also highlighted by the sparseness in the point cloud when compared to trajectories with good visibility. Table~\ref{table: APE results} summarizes the Absolute Position Error (APE) for the proposed method against ground truth considering a set of significant changes in the refractive index $n$, alongside the APE for a fixed model where the refractive terms are all calculated for $n=1.33$ as expected in such waters. Naturally, trajectories with good visibility lead to faster convergence, yet equally importantly the method converges even at low visibility conditions.

\subsection{Refractive Index Converge and Odometry ``in the Wild''}

\begin{figure}[ht!]
\centering
    \includegraphics[width=0.99\columnwidth]{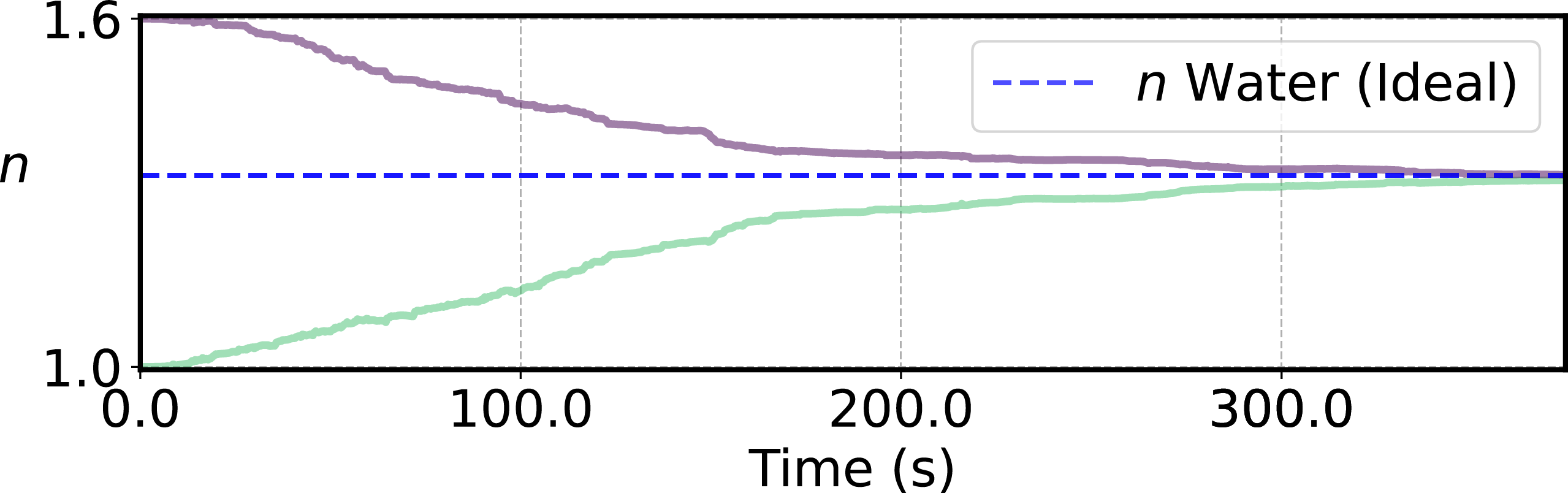}
\vspace{-2ex} 
\caption{Refractive index estimation convergence subject to ``wild'' initialization from $n$ equal to $1$ (i.e., that of air at STP) and $n$ equal to $1.6$ (e.g., value of carbon disulfide at $20^\circ$C). 
}
\label{fig:convergence_in_wild}
\end{figure}

Last but not least, we perform an experiment aiming to verify the ability of the proposed estimation solution to converge to the correct refractive index even after extremely wrong initialization. Two tests are conducted, namely with refractive index initially set to $1$ (air in Standard Temperature and Pressure (STP)) and $1.6$ (a value that for example is assumed by carbon disulfide at $20^\circ$C). The robot performs a trajectory involving a figure-8 maneuver. The results regarding the convergence of the refractive index are shown in Figure~\ref{fig:convergence_in_wild} and demonstrate the resilience of $n$ estimation.

\begin{table}
    \centering
    \caption{APE(Meters) for Estimated Odometry vs. Ground Truth} 
     \begin{tabular}{|c|c|c|c|c|c|c|c|}
    \hline
        \multicolumn{7}{|c|}{A.~~From t=0 (Complete trajectory)} \\ \hline
        \multicolumn{6}{|c|}{\textbf{Online - Initial} $n$} &  \textbf{Fixed} \\ \hline
        no. & \textbf{1.31} & \textbf{1.32} & \textbf{1.33} & \textbf{1.34} & \textbf{1.35} & \textbf{1.33} \\ \hline
        \textbf{1} & 0.6035 & 0.4419 & 0.4337 & 0.4936 & 0.4273 & 0.2962 \\ \hline
        \textbf{2} & 0.6363 & 0.5411 & 0.6413 & 0.7047 & 0.7332 & 0.4729 \\ \hline
        \textbf{3} & 0.6306 & 0.61 & 0.6022 & 0.578 & 0.609 & 0.5488 \\ \hline
        \textbf{4} & 0.6589 & 0.5483 & 0.5488 & 0.3963 & 0.5508 & 0.8611 \\ \hline
        \textbf{5} & 1.2654 & 1.1362 & 0.9941 & 0.9133 & 0.8226 & 1.0352 \\ \hline
        \textbf{6} & 0.5341 & 0.5312 & 0.6529 & 0.7791 & 0.8713 & 0.446 \\ \hline
        \hline
        \multicolumn{7}{|c|}{B.~~From t=150 } \\ \hline
        \multicolumn{6}{|c|}{\textbf{Online - Initial} $n$} &  \textbf{Fixed} \\ \hline
        no. & \textbf{1.31} & \textbf{1.32} & \textbf{1.33} & \textbf{1.34} & \textbf{1.35} & \textbf{1.33} \\ \hline
        \textbf{1} & 0.4112 & 0.3997 & 0.378  & 0.3701 & 0.3347 & 0.2947 \\ \hline
        \textbf{2} & 0.608  & 0.5782 & 0.6018 & 0.5983 & 0.625  & 0.4885 \\ \hline
        \textbf{3} & 0.4652 & 0.4264 & 0.4356 & 0.4465 & 0.4212 & 0.4504 \\ \hline
        \textbf{4} & 0.5082 & 0.57   & 0.4875 & 0.5352 & 0.4725 & 0.5008 \\ \hline
        \textbf{5} & 0.6171 & 0.7757 & 0.7669 & 0.7058 & 0.8364 & 1.0026 \\ \hline
        \textbf{6} & 0.3125 & 0.2776 & 0.2917 & 0.2254 & 0.2794 & 0.2794 \\ \hline
    \end{tabular}
    \label{table: APE results}
\end{table}

\section{Conclusions}\label{sec:concl}
This work presented a new general refractive camera model combined with an augmented visual-inertial fusion framework enabling co-estimation of refractive index and odometry. Verified through extensive experimental results, the method allows robust convergence of the refractive index even without a good initialization. Accordingly, it facilitates reliable visual-inertial odometry by only requiring conventional camera/\ac{imu} calibration in the air thus eliminating the need for the laborious task of medium-specific calibration or advanced knowledge of a medium's refractive index. 



\bibliographystyle{IEEEtran}
\bibliography{BIB/main,BIB/GeneralMonocularModel,BIB/DepthPrediction,BIB/CamModel_and_MVG,BIB/GeneralUnderwater,BIB/VIO}

\begin{thebibliography}{10}
\providecommand{\url}[1]{#1}
\csname url@samestyle\endcsname
\providecommand{\newblock}{\relax}
\providecommand{\bibinfo}[2]{#2}
\providecommand{\BIBentrySTDinterwordspacing}{\spaceskip=0pt\relax}
\providecommand{\BIBentryALTinterwordstretchfactor}{4}
\providecommand{\BIBentryALTinterwordspacing}{\spaceskip=\fontdimen2\font plus
\BIBentryALTinterwordstretchfactor\fontdimen3\font minus \fontdimen4\font\relax}
\providecommand{\BIBforeignlanguage}[2]{{%
\expandafter\ifx\csname l@#1\endcsname\relax
\typeout{** WARNING: IEEEtran.bst: No hyphenation pattern has been}%
\typeout{** loaded for the language `#1'. Using the pattern for}%
\typeout{** the default language instead.}%
\else
\language=\csname l@#1\endcsname
\fi
#2}}
\providecommand{\BIBdecl}{\relax}
\BIBdecl

\bibitem{schill2018vertex}
F.~Schill, A.~Bahr, and A.~Martinoli, ``Vertex: A new distributed underwater robotic platform for environmental monitoring,'' in \emph{Distributed Autonomous Robotic Systems: The 13th International Symposium}.\hskip 1em plus 0.5em minus 0.4em\relax Springer, 2018, pp. 679--693.

\bibitem{galceran2013planning}
E.~Galceran and M.~Carreras, ``Planning coverage paths on bathymetric maps for in-detail inspection of the ocean floor,'' in \emph{2013 IEEE International Conference on Robotics and Automation}.\hskip 1em plus 0.5em minus 0.4em\relax IEEE, 2013, pp. 4159--4164.

\bibitem{hollinger2013active}
G.~A. Hollinger, B.~Englot, F.~S. Hover, U.~Mitra, and G.~S. Sukhatme, ``Active planning for underwater inspection and the benefit of adaptivity,'' \emph{The International Journal of Robotics Research}, vol.~32, no.~1, pp. 3--18, 2013.

\bibitem{wu2019survey}
Y.~Wu, X.~Ta, R.~Xiao, Y.~Wei, D.~An, and D.~Li, ``Survey of underwater robot positioning navigation,'' \emph{Applied Ocean Research}, vol.~90, p. 101845, 2019.

\bibitem{bahr2009cooperative}
A.~Bahr, J.~J. Leonard, and M.~F. Fallon, ``Cooperative localization for autonomous underwater vehicles,'' \emph{The International Journal of Robotics Research}, vol.~28, no.~6, pp. 714--728, 2009.

\bibitem{paull2013auv}
L.~Paull, S.~Saeedi, M.~Seto, and H.~Li, ``Auv navigation and localization: A review,'' \emph{IEEE Journal of oceanic engineering}, vol.~39, no.~1, pp. 131--149, 2013.

\bibitem{xu2022robust}
Y.~Xu, R.~Zheng, S.~Zhang, and M.~Liu, ``Robust inertial-aided underwater localization based on imaging sonar keyframes,'' \emph{IEEE Transactions on Instrumentation and Measurement}, vol.~71, pp. 1--12, 2022.

\bibitem{johannsson2010imaging}
H.~Johannsson, M.~Kaess, B.~Englot, F.~Hover, and J.~Leonard, ``Imaging sonar-aided navigation for autonomous underwater harbor surveillance,'' in \emph{2010 IEEE/RSJ International Conference on Intelligent Robots and Systems}.\hskip 1em plus 0.5em minus 0.4em\relax IEEE, 2010, pp. 4396--4403.

\bibitem{shukla2016application}
A.~Shukla and H.~Karki, ``Application of robotics in offshore oil and gas industry—a review part ii,'' \emph{Robotics and Autonomous Systems}, vol.~75, pp. 508--524, 2016.

\bibitem{ferrera2019aqualoc}
M.~Ferrera, V.~Creuze, J.~Moras, and P.~Trouv{\'e}-Peloux, ``Aqualoc: An underwater dataset for visual--inertial--pressure localization,'' \emph{The International Journal of Robotics Research}, vol.~38, no.~14, pp. 1549--1559, 2019.

\bibitem{miao2021univio}
R.~Miao, J.~Qian, Y.~Song, R.~Ying, and P.~Liu, ``Univio: Unified direct and feature-based underwater stereo visual-inertial odometry,'' \emph{IEEE Transactions on Instrumentation and Measurement}, vol.~71, pp. 1--14, 2021.

\bibitem{teixeira2020deep}
B.~Teixeira, H.~Silva, A.~Matos, and E.~Silva, ``Deep learning for underwater visual odometry estimation,'' \emph{IEEE Access}, vol.~8, pp. 44\,687--44\,701, 2020.

\bibitem{rahman2019svin2}
S.~Rahman, A.~Q. Li, and I.~Rekleitis, ``Svin2: An underwater slam system using sonar, visual, inertial, and depth sensor,'' in \emph{2019 IEEE/RSJ International Conference on Intelligent Robots and Systems (IROS)}.\hskip 1em plus 0.5em minus 0.4em\relax IEEE, 2019, pp. 1861--1868.

\bibitem{randall2023flsea}
Y.~Randall, ``Flsea: Underwater visual-inertial and stereo-vision forward-looking datasets,'' Ph.D. dissertation, University of Haifa (Israel), 2023.

\bibitem{joshi2023sm}
B.~Joshi, H.~Damron, S.~Rahman, and I.~Rekleitis, ``Sm/vio: Robust underwater state estimation switching between model-based and visual inertial odometry,'' \emph{arXiv preprint arXiv:2304.01988}, 2023.

\bibitem{gu2019environment}
C.~Gu, Y.~Cong, and G.~Sun, ``Environment driven underwater camera-imu calibration for monocular visual-inertial slam,'' in \emph{2019 International Conference on Robotics and Automation (ICRA)}.\hskip 1em plus 0.5em minus 0.4em\relax IEEE, 2019, pp. 2405--2411.

\bibitem{austinIndexRefractionSeawater1976}
\BIBentryALTinterwordspacing
R.~W. Austin and G.~Halikas, ``The index of refraction of seawater.'' [Online]. Available: \url{https://escholarship.org/uc/item/8px2019m}
\BIBentrySTDinterwordspacing

\bibitem{singhOnlineSelfcalibratingRefractive2023}
M.~Singh, M.~Dharmadhikari, and K.~Alexis, ``An online self-calibrating refractive camera model with application to underwater odometry,'' in \emph{2024 IEEE International Conference on Robotics and Automation (ICRA)}, 2024, pp. 10\,005--10\,011.

\bibitem{hartley2003multiple}
R.~Hartley and A.~Zisserman, \emph{Multiple view geometry in computer vision}.\hskip 1em plus 0.5em minus 0.4em\relax Cambridge university press, 2003.

\bibitem{fitzgibbon2001simultaneous}
A.~W. Fitzgibbon, ``Simultaneous linear estimation of multiple view geometry and lens distortion,'' in \emph{Proceedings of the 2001 IEEE Computer Society Conference on Computer Vision and Pattern Recognition. CVPR 2001}, vol.~1.\hskip 1em plus 0.5em minus 0.4em\relax IEEE, 2001, pp. I--I.

\bibitem{barreto2005fundamental}
J.~P. Barreto and K.~Daniilidis, ``Fundamental matrix for cameras with radial distortion,'' in \emph{Tenth IEEE International Conference on Computer Vision (ICCV'05) Volume 1}, vol.~1.\hskip 1em plus 0.5em minus 0.4em\relax IEEE, 2005, pp. 625--632.

\bibitem{willson1994center}
R.~G. Willson and S.~A. Shafer, ``What is the center of the image?'' \emph{JOSA A}, vol.~11, no.~11, pp. 2946--2955, 1994.

\bibitem{chari2009multiple}
V.~Chari and P.~Sturm, ``Multiple-view geometry of the refractive plane,'' in \emph{BMVC 2009-20th British machine vision conference}.\hskip 1em plus 0.5em minus 0.4em\relax The British Machine Vision Association (BMVA), 2009, pp. 1--11.

\bibitem{huang2017plate}
L.~Huang, X.~Zhao, S.~Cai, and Y.~Liu, ``Plate refractive camera model and its applications,'' \emph{Journal of Electronic Imaging}, vol.~26, no.~2, pp. 023\,020--023\,020, 2017.

\bibitem{treibitz2011flat}
T.~Treibitz, Y.~Schechner, C.~Kunz, and H.~Singh, ``Flat refractive geometry,'' \emph{IEEE transactions on pattern analysis and machine intelligence}, vol.~34, no.~1, pp. 51--65, 2011.

\bibitem{sedlazeck2012perspective}
A.~Sedlazeck and R.~Koch, ``Perspective and non-perspective camera models in underwater imaging--overview and error analysis,'' in \emph{Outdoor and Large-Scale Real-World Scene Analysis: 15th International Workshop on Theoretical Foundations of Computer Vision, Dagstuhl Castle, Germany, June 26-July 1, 2011. Revised Selected Papers}.\hskip 1em plus 0.5em minus 0.4em\relax Springer, 2012, pp. 212--242.

\bibitem{agrawal2012theory}
A.~Agrawal, S.~Ramalingam, Y.~Taguchi, and V.~Chari, ``A theory of multi-layer flat refractive geometry,'' in \emph{2012 IEEE conference on computer vision and pattern recognition}.\hskip 1em plus 0.5em minus 0.4em\relax IEEE, 2012, pp. 3346--3353.

\bibitem{jordt2013refractive}
A.~Jordt-Sedlazeck and R.~Koch, ``Refractive structure-from-motion on underwater images,'' in \emph{Proceedings of the IEEE international Conference on Computer Vision}, 2013, pp. 57--64.

\bibitem{kawahara2013pixel}
R.~Kawahara, S.~Nobuhara, and T.~Matsuyama, ``A pixel-wise varifocal camera model for efficient forward projection and linear extrinsic calibration of underwater cameras with flat housings,'' in \emph{Proceedings of the IEEE International Conference on Computer Vision Workshops}, 2013, pp. 819--824.

\bibitem{chaudhury2015multiple}
S.~Chaudhury, T.~Agarwal, and P.~Maheshwari, ``Multiple view 3-d reconstruction in water,'' in \emph{2015 Fifth National Conference on Computer Vision, Pattern Recognition, Image Processing and Graphics (NCVPRIPG)}.\hskip 1em plus 0.5em minus 0.4em\relax IEEE, 2015, pp. 1--4.

\bibitem{haner2015absolute}
S.~Haner and K.~Astrom, ``Absolute pose for cameras under flat refractive interfaces,'' in \emph{Proceedings of the IEEE conference on computer vision and pattern recognition}, 2015, pp. 1428--1436.

\bibitem{hu2023refractive}
X.~Hu, F.~Lauze, and K.~S. Pedersen, ``Refractive pose refinement: Generalising the geometric relation between camera and refractive interface,'' \emph{International Journal of Computer Vision}, vol. 131, no.~6, pp. 1448--1476, 2023.

\bibitem{hu2021absolute}
X.~Hu, F.~Lauze, K.~S. Pedersen, and J.~M{\'e}lou, ``Absolute and relative pose estimation in refractive multi view,'' in \emph{Proceedings of the IEEE/CVF international conference on computer vision}, 2021, pp. 2569--2578.

\bibitem{zhang2021open}
P.~Zhang, Z.~Wu, J.~Wang, S.~Kong, M.~Tan, and J.~Yu, ``An open-source, fiducial-based, underwater stereo visual-inertial localization method with refraction correction,'' in \emph{2021 IEEE/RSJ International Conference on Intelligent Robots and Systems (IROS)}.\hskip 1em plus 0.5em minus 0.4em\relax IEEE, 2021, pp. 4331--4336.

\bibitem{servos2013underwater}
J.~Servos, M.~Smart, and S.~L. Waslander, ``Underwater stereo slam with refraction correction,'' in \emph{2013 IEEE/RSJ International Conference on Intelligent Robots and Systems}.\hskip 1em plus 0.5em minus 0.4em\relax IEEE, 2013, pp. 3350--3355.

\bibitem{suresh2019through}
S.~Suresh, E.~Westman, and M.~Kaess, ``Through-water stereo slam with refraction correction for auv localization,'' \emph{IEEE Robotics and Automation Letters}, vol.~4, no.~2, pp. 692--699, 2019.

\bibitem{shkurti2011state}
F.~Shkurti, I.~Rekleitis, M.~Scaccia, and G.~Dudek, ``State estimation of an underwater robot using visual and inertial information,'' in \emph{2011 IEEE/RSJ International Conference on Intelligent Robots and Systems}.\hskip 1em plus 0.5em minus 0.4em\relax IEEE, 2011, pp. 5054--5060.

\bibitem{rahman2018sonar}
S.~Rahman, A.~Q. Li, and I.~Rekleitis, ``Sonar visual inertial slam of underwater structures,'' in \emph{2018 IEEE International Conference on Robotics and Automation (ICRA)}.\hskip 1em plus 0.5em minus 0.4em\relax IEEE, 2018, pp. 5190--5196.

\bibitem{hu2022tightly}
C.~Hu, S.~Zhu, Y.~Liang, and W.~Song, ``Tightly-coupled visual-inertial-pressure fusion using forward and backward imu preintegration,'' \emph{IEEE Robotics and Automation Letters}, vol.~7, no.~3, pp. 6790--6797, 2022.

\bibitem{bloeschIteratedExtendedKalman2017}
\BIBentryALTinterwordspacing
M.~Bloesch, M.~Burri, S.~Omari, M.~Hutter, and R.~Siegwart, ``Iterated extended {{Kalman}} filter based visual-inertial odometry using direct photometric feedback,'' vol.~36, no.~10, pp. 1053--1072. [Online]. Available: \url{https://doi.org/10.1177/0278364917728574}
\BIBentrySTDinterwordspacing

\bibitem{kannalaGenericCameraModel2006}
J.~Kannala and S.~Brandt, ``A {{Generic Camera Model}} and {{Calibration Method}} for {{Conventional}}, {{Wide-Angle}}, and {{Fish-Eye Lenses}},'' vol.~28, pp. 1335--40.

\bibitem{cerberus_science}
M.~Tranzatto, T.~Miki, M.~Dharmadhikari, L.~Bernreiter, M.~Kulkarni, F.~Mascarich, O.~Andersson, S.~Khattak, M.~Hutter, R.~Siegwart \emph{et~al.}, ``Cerberus in the darpa subterranean challenge,'' \emph{Science Robotics}, vol.~7, no.~66, p. eabp9742, 2022.

\bibitem{cerberus_finals}
M.~Tranzatto, M.~Dharmadhikari, L.~Bernreiter, M.~Camurri, S.~Khattak, F.~Mascarich, P.~Pfreundschuh, D.~Wisth, S.~Zimmermann, M.~Kulkarni \emph{et~al.}, ``Team cerberus wins the darpa subterranean challenge: Technical overview and lessons learned,'' \emph{arXiv preprint arXiv:2207.04914}, 2022.

\bibitem{cerberus_tunnel_urbam}
M.~Tranzatto, F.~Mascarich, L.~Bernreiter, C.~Godinho, M.~Camurri, S.~Khattak, T.~Dang, V.~Reijgwart, J.~Loeje, D.~Wisth \emph{et~al.}, ``Cerberus: Autonomous legged and aerial robotic exploration in the tunnel and urban circuits of the darpa subterranean challenge,'' \emph{arXiv preprint arXiv:2201.07067}, 2022.

\bibitem{daimon2007measurement}
M.~Daimon and A.~Masumura, ``Measurement of the refractive index of distilled water from the near-infrared region to the ultraviolet region,'' \emph{Applied optics}, vol.~46, no.~18, pp. 3811--3820, 2007.

\bibitem{lide2004crc}
D.~R. Lide, \emph{CRC handbook of chemistry and physics}.\hskip 1em plus 0.5em minus 0.4em\relax CRC press, 2004, vol.~85.

\bibitem{lipson2021raftstereo}
L.~Lipson, Z.~Teed, and J.~Deng, ``Raft-stereo: Multilevel recurrent field transforms for stereo matching,'' in \emph{International Conference on 3D Vision (3DV)}, 2021.

\end{thebibliography}

\end{document}